\documentclass[conference]{IEEEtran}
\IEEEoverridecommandlockouts
\usepackage{cite}
\usepackage{amsmath,amssymb,amsfonts}
\usepackage{algorithm}
\usepackage{algpseudocode}
\usepackage{graphicx}
\usepackage{textcomp}
\usepackage{xcolor}
\usepackage{multirow}
\usepackage{booktabs}
\usepackage{setspace}
\usepackage{comment}
\usepackage{enumitem}
\usepackage{setspace}
\usepackage{caption}
\usepackage{url}
\usepackage{subcaption}
\newlist{inlineroman}{enumerate*}{1}
\setlist[inlineroman]{itemjoin*={{ }},afterlabel=~,label=\roman*.}

\usepackage[compatibility=false]{caption}

\renewcommand{\baselinestretch}{0.97}

\usepackage{makecell}
\algblock{Input}{EndInput}
\algnotext{EndInput}
\algblock{Ensure}{EndEnsure}
\algnotext{EndEnsure}
\algblock{Output}{EndOutput}
\algnotext{EndOutput}

\def\BibTeX{{\rm B\kern-.05em{\sc i\kern-.025em b}\kern-.08em
    T\kern-.1667em\lower.7ex\hbox{E}\kern-.125emX}}

\begin{document}

\title{
GraphLeap: Decoupling Graph Construction and Convolution for Vision GNN Acceleration on FPGA}



\author{
\IEEEauthorblockN{Anvitha Ramachandran, Dhruv Parikh, Viktor Prasanna}
\IEEEauthorblockA{
  University of Southern California, Los Angeles, California, USA \\
   alramach@usc.edu, dhruvash@usc.edu, prasanna@usc.edu}
}

\maketitle

\begin{abstract}
Vision Graph Neural Networks (ViGs) model an image as a graph of patch tokens, enabling adaptive, feature-driven neighborhoods. Unlike CNNs with fixed grid biases or Vision Transformers with global token interactions, ViGs rely on dynamic graph convolution: at each layer, a feature-dependent graph is built via $k$-nearest neighbor (kNN) search on current patch features, followed by message passing. This per-layer graph construction is the main bottleneck—consuming 50–95\% of graph convolution time on CPUs and GPUs—and scales by $O(N^2)$ where $N$ is the number of patches, creating a sequential dependency between graph construction and feature updates.

In this paper, we introduce \textit{GraphLeap}, a simple yet novel reformulation that removes this dependency by decoupling graph construction from feature update across layers. GraphLeap performs the feature update at layer $\ell$ using a graph constructed from the previous layer's features, while simultaneously using the current layer's features to construct the graph for layer $\ell{+}1$. This one-layer lookahead dynamic graph construction enables concurrent graph construction and GNN message passing, yielding a new class of ViGs, which we call \textit{GraphLeap}. While using prior-layer features for graph construction can introduce minor accuracy degradation, we show that lightweight fine-tuning for a few epochs is sufficient to recover original accuracy. Building on \textit{GraphLeap}, we present the first end-to-end FPGA accelerator for Vision GNNs. Our design is a streaming, layer-pipelined architecture that overlaps a kNN graph construction engine with a feature update engine, exploits node and channel-level parallelism, and enables efficient on-chip dataflow without explicit materialization of edge features. Evaluated on isotropic and pyramidal ViG models deployed on an Alveo U280 FPGA, our approach achieves up to $95.7\times$ speedup over CPU and $8.5\times$ speedup over GPU baselines, demonstrating the feasibility of real-time Vision GNN inference and motivating future hardware--algorithm co-design for graph-based vision models.
Code is available at \url{https://github.com/anvitha305/GraphLeap}.



\end{abstract}

\begin{IEEEkeywords}
Graph Neural Networks, Vision GNNs, image graph construction, hardware acceleration
\end{IEEEkeywords}

\section{Introduction}
\label{sec:intro}

Modern computer vision backbones span Convolutional Neural Networks (CNNs), Vision Transformers (ViTs), state-space based vision models, and Vision Graph Neural Networks (ViGs). 
CNNs apply convolutional filters on grid-structured feature maps to exploit locality \cite{CNNs, AlexNet}. 
ViTs treat image patches as tokens and use self-attention to capture long-range interactions, at the cost of quadratic complexity in the number of tokens \cite{VisionTransformers, raghu2022visiontransformerslikeconvolutional}. 
More recently, state-space based vision models (e.g., Vision Mamba) trade explicit pairwise attention for selective scan style sequence modeling \cite{Mamba, VisionMamba}. 
ViGs depart from grids and sequences by representing an image as a graph of patch tokens whose edges encode feature-driven patch relationships; critically, this graph is reconstructed across layers to adapt neighborhoods to evolving features \cite{VisionGNN}. 
This flexibility allows ViGs to achieve competitive accuracy with favorable compute efficiency compared to similarly sized ViTs, making them suitable for resource-constrained settings \cite{MobileViG, ViHGNN}.

Dynamic graph construction remains a major bottleneck for real-time ViG deployment. Prior methods reduce its cost via clustering, windowing, or approximate neighborhoods \cite{ClusterViG, WiGNet, MobileViG2, AdaptiveViG, AttentionViG, SearchViG, MobileViG, ViHGNN, DVHGNN, PVG, EvGNN, GreedyVIG}, but still enforce per-layer dependencies and often compromise the dynamic, feature-driven graphs central to ViGs. Most GNN accelerators target large static graphs, while vision accelerators focus on CNN/ViT/state-space operators, leaving dynamic image graph construction and full ViG pipelines largely unsupported.

Motivated by these limitations, we ask: \emph{can dynamic graph construction be decoupled from graph convolution in ViGs?} We introduce \textit{GraphLeap}, which updates features at layer $\ell$ using the previous layer’s graph while concurrently constructing the graph for layer $\ell+1$ from current features. This one-layer lookahead enables overlapping graph construction and convolution, reducing inference latency. We show that any minor accuracy loss from using prior-layer graphs can be recovered with lightweight fine-tuning.

Building on GraphLeap, we present the first end-to-end FPGA accelerator for Vision GNN inference. 
Our design is a streaming, layer-pipelined architecture that overlaps a kNN-based graph construction engine with a feature update engine, exploits node- and channel-level parallelism, and supports ViG graph convolution efficiently in hardware. 
Our contributions are:
\begin{itemize}
    \item \textbf{GraphLeap Vision GNNs.} We propose GraphLeap, a general algorithmic reformulation that decouples dynamic graph construction from graph convolution by constructing the graph for layer $\ell{+}1$ while updating features for layer $\ell$, and we validate its effectiveness across standard isotropic and pyramidal ViG backbones.
    \item \textbf{End-to-end FPGA accelerator.} We develop the first end-to-end FPGA accelerator for ViG inference based on GraphLeap, implementing a streaming pipeline that overlaps graph construction and feature update across layers to reduce end-to-end latency.
    \item \textbf{Streaming kNN graph construction.} We design a high-throughput dynamic graph construction engine for dilated kNN image graphs that concurrently operates one layer ahead of feature update and streams neighbor indices to the compute pipeline.
    \item \textbf{Efficient feature-update pipeline.} We design a hardware feature-update engine specialized for ViGs, including a shared systolic MLP compute fabric reused across graph convolution and Feed Forward Network (FFN) stages via streaming buffers.
    \item \textbf{Evaluation.} We demonstrate the effectiveness of the \textit{GraphLeap} architecture on an AMD-Xilinx Alveo U280 FPGA across various isotropic and pyramidal ViG variants. Our results show significant end-to-end performance gains, achieving up to \textbf{95.7$\times$} speedup over CPU and \textbf{8.5$\times$} over GPU. 
    \item \textbf{Comparison to ViT Accelerators.} We compare the latency of our accelerator with recent ViT accelerators. Compared to state-of-the-art ViT accelerators on similar FPGA platforms, our accelerator achieves an end-to-end latency that is $3\times$ faster with a comparable accuracy.
\end{itemize}

\section{Challenges and Key Innovations}
\label{sec: chall}
\subsection{Preliminaries: Graph Neural Networks}
\label{sec:prelim_gnn}

\paragraph{Graph Neural Networks} A graph is denoted as $\mathcal{G}=(\mathcal{V},\mathcal{E})$, where $\mathcal{V}$ is the node set and $\mathcal{E}$ is the edge set. Let $N = |\mathcal{V}|$ and node features be stacked as $\mathbf{X}\in\mathbb{R}^{N\times D}$, where $\mathbf{x}_i\in\mathbb{R}^{D}$ is the feature of node $v_i$. Most GNNs can be expressed in the message-passing form: each layer aggregates neighbor information and then updates node states (features) \cite{gori2005new, scarselli2008graph, gilmer2017neural, kipf2016semi, hamilton2017inductive}.
Concretely, with neighbor set $\mathcal{N}(i)$ (induced by $\mathcal{E}$), a generic layer is
\begin{equation}
\mathbf{x}_i' = \mathrm{Upd}\!\Big(\mathbf{x}_i,\ \mathrm{Agg}\big(\{\mathbf{x}_j : j\in\mathcal{N}(i)\}\big)\Big),
\end{equation}
where $\mathrm{Agg}(\cdot)$ is a permutation-invariant operator (e.g., sum/mean/max), and $\mathrm{Upd}(\cdot)$ is typically a linear layer or an MLP.

\paragraph{Max-Relative Graph Convolution (MRConv)} Following \cite{deepgcn, VisionGNN}, we focus on max-relative aggregation, which is simple and efficient:
\begin{align}
\mathbf{x}_i'' &= \Big[\,\mathbf{x}_i,\ \max_{j\in\mathcal{N}(i)}(\mathbf{x}_j-\mathbf{x}_i)\,\Big] \in \mathbb{R}^{2D}, \label{eq:mrconv_agg}\\
\mathbf{x}_i'  &= \mathbf{x}_i'' \mathbf{W}_{\mathrm{upd}}, \label{eq:mrconv_upd}
\end{align}
where $\mathbf{W}_{\mathrm{upd}}\in\mathbb{R}^{2D\times D}$ (bias omitted). In practice, ViG further adopts a multi-head update: split $\mathbf{x}_i''$ into $h$ heads and apply per-head weights, then concatenate the results \cite{VisionGNN}.

\subsection{ViG: Dynamic Graph Convolution over Image Patches}
\label{sec:prelim_vig}

\paragraph{Image-to-graph representation} ViG represents an image as a graph over non-overlapping patches \cite{VisionGNN}. Let an image be divided into $N$ patches, each mapped to a node feature, yielding $\mathbf{X}\in\mathbb{R}^{N\times D}$. A \emph{dynamic} image graph is constructed by connecting each node to its $K$ nearest neighbors in feature space:
\begin{equation}
\mathcal{G} = G(\mathbf{X}),\qquad \mathcal{G}=(\mathcal{V},\mathcal{E}),\quad \mathcal{N}(i)\ \text{derived from}\ \mathcal{E}.
\end{equation}
Unlike typical GNN settings with static graphs, ViG \emph{reconstructs} the graph at each layer using the current features, enabling content-adaptive neighborhoods \cite{VisionGNN}.

\paragraph{The ViG block with Grapher + FFN} A ViG model is built by stacking repeated blocks. Given $\mathbf{X}^{(\ell)}\in\mathbb{R}^{N\times D}$ as the input to block $\ell$:

\noindent\textbf{Grapher.}
In the standard ViG formulation, the graph $\mathcal{G}^{(\ell)}$ is constructed from the projected features $\mathbf{U}^{(\ell)}$ of the \textit{current} layer:
\begin{align}
\mathbf{U}^{(\ell)} &= \mathbf{X}^{(\ell)} \mathbf{W}_{\mathrm{in}}^{(\ell)}, \label{eq:vig_fc1}\\
\mathcal{G}^{(\ell)} &= G\!\left(\mathbf{U}^{(\ell)}\right), \label{eq:vig_graph}\\
\widetilde{\mathbf{U}}^{(\ell)} &= \mathrm{MRConv}\!\left(\mathbf{U}^{(\ell)};\ \mathcal{G}^{(\ell)}\right), \label{eq:vig_mrconv}\\
\mathbf{Y}^{(\ell)} &= \sigma\!\left(\widetilde{\mathbf{U}}^{(\ell)}\right)\mathbf{W}_{\mathrm{out}}^{(\ell)} + \mathbf{X}^{(\ell)}. \label{eq:vig_grapher}
\end{align}

\noindent\textbf{FFN.}
The FFN then updates the features for the next layer:
\begin{equation}
\mathbf{X}^{(\ell+1)} = \sigma\!\left(\mathbf{Y}^{(\ell)}\mathbf{W}_1^{(\ell)}\right)\mathbf{W}_2^{(\ell)} + \mathbf{Y}^{(\ell)}. \label{eq:vig_ffn}
\end{equation}

\subsection{Challenges and Key Innovations}

\noindent\textbf{Challenge \#1: Dynamic graph construction is both expensive and sequential.}
Vision GNNs differ from traditional GNNs because the graph is \emph{reconstructed at every layer}. This $k$NN construction accounts for 50--95\% of total inference time and creates a strict sequential dependency: graph construction $\rightarrow$ convolution. In a standard ViG layer, the graph construction ($\mathcal{G}^{(\ell)}$) must complete before the feature update ($\mathbf{X}^{(\ell+1)}$) can begin, enforcing a serialized execution that dominates end-to-end latency.

\noindent\textbf{Key Innovation (GraphLeap):} We decouple this bottleneck by reformulating the layer dependency. In GraphLeap, the feature update at layer $\ell$ utilizes a "leap-ahead" graph constructed from the features of the \textit{previous} layer:
\begin{equation}
    \mathcal{G}^{(\ell)} = G(\mathbf{X}^{(\ell-1)}).
\end{equation}

This allows the Graph Construction Engine (GCE) to build the graph for layer $\ell+1$ using $\mathbf{X}^{(\ell)}$ while the Feature Update Engine (FUE) concurrently processes layer $\ell$. We demonstrate that any accuracy loss from this shifted connectivity is recovered with lightweight fine-tuning.

\noindent\textbf{Challenge \#2: Mixing irregular graph kernels with regular dense MLP compute.}
ViG blocks couple irregular kernels (distance evaluation, top-$k$ selection) with regular dense operations (FFNs). Prior accelerators focus on static graphs or regular grids, leaving dynamic image graph pipelines unsupported. 
\textit{Key Innovation:} Our design treats ViG as an end-to-end streaming workload and co-designs a dedicated $k$NN pipeline with a dense systolic MLP datapath, connected via FIFOs to maintain throughput under mixed behavior.

\noindent\textbf{Challenge \#3: Intermediate graph/edge materialization is prohibitive.}
Naively materializing edge features scales as $O(NK)$ and exhausts on-chip memory at ViG resolutions. 
\textit{Key Innovation:} We stream neighbor indices and features through the pipeline and compute aggregation on-the-fly, keeping only node features and small line buffers on-chip, avoiding explicit $O(NK)$ edge-feature storage.

\noindent\textbf{Challenge \#4: ViG families introduce shape variation across layers.}
Pyramid ViGs change patch resolution ($N$) and feature widths ($D$) across stages. 
\textit{Key Innovation:} We parameterize tiling and parallelization along both the node and channel dimensions. We enforce clean stage boundaries where the GCE and FUE resynchronize, preserving the GraphLeap overlap while supporting both isotropic and pyramidal models.
\section{GraphLeap}
\label{sec: ddvig}

\subsection{GraphLeap: Decoupling Graph Construction and Graph Convolution}
\label{sec:graphleap}

\paragraph{Core idea of leap-ahead graph construction}
In standard ViG, each block $\ell$ must first build $\mathcal{G}^{(\ell)}$ from the current projected features $\mathbf{U}^{(\ell)}$ (Eq.~\ref{eq:vig_graph}) and only then apply MRConv using that graph (Eq.~\ref{eq:vig_mrconv}). GraphLeap breaks this strict order by allowing the graph used at block $\ell$ to be constructed from \emph{previous} projected features, while the \emph{current} projected features are simultaneously used to construct the \emph{next} graph.

We denote by $\widehat{\mathcal{G}}^{(\ell)}$ the graph actually used for MRConv at block $\ell$ under GraphLeap. GraphLeap defines:
\begin{align}
\mathbf{U}^{(\ell)} &= \mathbf{X}^{(\ell)} \mathbf{W}_{\mathrm{in}}^{(\ell)}, \label{eq:gl_fc1}\\
\widehat{\mathcal{G}}^{(\ell)} &= G\!\left(\mathbf{U}^{(\ell-1)}\right)\quad (\ell\ge 1), \label{eq:gl_graph_prev}\\
\widetilde{\mathbf{U}}^{(\ell)} &= \mathrm{MRConv}\!\left(\mathbf{U}^{(\ell)};\ \widehat{\mathcal{G}}^{(\ell)}\right), \label{eq:gl_mrconv}\\
\mathbf{Y}^{(\ell)} &= \sigma\!\left(\widetilde{\mathbf{U}}^{(\ell)}\right)\mathbf{W}_{\mathrm{out}}^{(\ell)} + \mathbf{X}^{(\ell)}, \label{eq:gl_grapher}\\
\mathbf{X}^{(\ell+1)} &= \sigma\!\left(\mathbf{Y}^{(\ell)}\mathbf{W}_1^{(\ell)}\right)\mathbf{W}_2^{(\ell)} + \mathbf{Y}^{(\ell)}. \label{eq:gl_ffn}
\end{align}
For the first block, we use the standard construction $\widehat{\mathcal{G}}^{(0)} = G(\mathbf{U}^{(0)})$ to bootstrap.

\paragraph{Classification as a ``Dynamic`` Vision GNN Architecture} GraphLeap does not replace $k$NN with a static rule; it still constructs $\widehat{\mathcal{G}}^{(\ell)}$ from features, but shifts \emph{which} features are used: the graph used at block $\ell$ is computed from $\mathbf{U}^{(\ell-1)}$ rather than $\mathbf{U}^{(\ell)}$. Intuitively, consecutive ViG blocks transform features smoothly (due to residual structure in Eq.~\ref{eq:vig_grapher}--\ref{eq:vig_ffn}), so a one-block ``staleness'' in the graph often preserves neighborhood quality while enabling concurrent execution of (i) building $G(\mathbf{U}^{(\ell)})$ and (ii) running MRConv using $G(\mathbf{U}^{(\ell-1)})$.

\paragraph{Training note} Using $\widehat{\mathcal{G}}^{(\ell)} = G(\mathbf{U}^{(\ell-1)})$ can introduce a small accuracy drop if applied naively. In our experiments, lightweight fine-tuning recovers this gap and can match or exceed the baseline ViG accuracy. This suggests that the model can adapt its representations so that the leap-ahead graphs remain effective for message passing.

\paragraph{Algorithm} Algorithm~\ref{alg:graphleap} summarizes GraphLeap at the level of the ViG block. It mirrors the standard ViG computation \cite{VisionGNN} while changing only the source of the graph used by MRConv.

\begin{algorithm}
\caption{GraphLeap ViG Block}
\label{alg:graphleap}
\begin{algorithmic}[1]
\Require Block input $\mathbf{X}^{(\ell)}\in\mathbb{R}^{N\times D}$; prev. features $\mathbf{U}^{(\ell-1)}$ (for $\ell{=}0$, $\mathbf{U}^{(-1)}\leftarrow \mathbf{X}^{(0)}\mathbf{W}_{\mathrm{in}}^{(0)}$).
\Statex \vspace{-2pt} 
\State $\mathbf{U}^{(\ell)} \leftarrow \mathbf{X}^{(\ell)} \mathbf{W}_{\mathrm{in}}^{(\ell)}$ \hfill \Comment{FC1}
\State $\widehat{\mathcal{G}}^{(\ell)} \leftarrow G\!\left(\mathbf{U}^{(\ell-1)}\right)$ \hfill \Comment{leap-ahead graph for this block}
\State $\widetilde{\mathbf{U}}^{(\ell)} \leftarrow \mathrm{MRConv}\!\left(\mathbf{U}^{(\ell)};\widehat{\mathcal{G}}^{(\ell)}\right)$ \hfill \Comment{message passing}
\State $\mathbf{Y}^{(\ell)} \leftarrow \sigma(\widetilde{\mathbf{U}}^{(\ell)})\mathbf{W}_{\mathrm{out}}^{(\ell)} + \mathbf{X}^{(\ell)}$ \hfill \Comment{FC2 + residual}
\State $\mathbf{X}^{(\ell+1)} \leftarrow \sigma(\mathbf{Y}^{(\ell)}\mathbf{W}_1^{(\ell)})\mathbf{W}_2^{(\ell)} + \mathbf{Y}^{(\ell)}$ \hfill \Comment{FFN + residual}
\State \Return $\mathbf{X}^{(\ell+1)}$ and cache $\mathbf{U}^{(\ell)}$ for constructing $\widehat{\mathcal{G}}^{(\ell+1)}$
\Statex \vspace{-4pt} 
\end{algorithmic}
\end{algorithm}
\section{System Architecture}
\label{sec:architecture}

In this section, we introduce the architecture of the end-to-end accelerator for ViG. We describe the \textbf{Graph Construction Engine (GCE)} for dilated $k$-NN graph construction and the \textbf{Feature Update Engine (FUE)} which implements the ViG \textbf{Grapher} and \textbf{FFN} modules.

\subsection{Accelerator Overview}
Our accelerator employs a dual-stage pipelined architecture that overlaps graph construction for layer $\ell+1$ with feature updates for layer $\ell$. The design consists of a Graph Construction Engine (GCE) and a Feature Update Engine (FUE), which operate concurrently and communicate via streaming interfaces.

The GCE generates graph connectivity directly from per-layer features and streams edge information without materializing full adjacency structures in off-chip memory, using a look-ahead buffer to decouple production and consumption. The FUE consumes this streamed data and performs feature updates through three integrated modules: the Gather Module (GM), which organizes irregular neighbor accesses; the Graph Convolution Module (GCM), which performs aggregation and transformation; and the Feedforward Module (FM), which applies per-node updates. Execution is scheduled such that the GCE remains one layer ahead of the FUE, ensuring that connectivity data is available when needed.
\begin{figure}[t]
\centering
\includegraphics[width=0.48\textwidth]{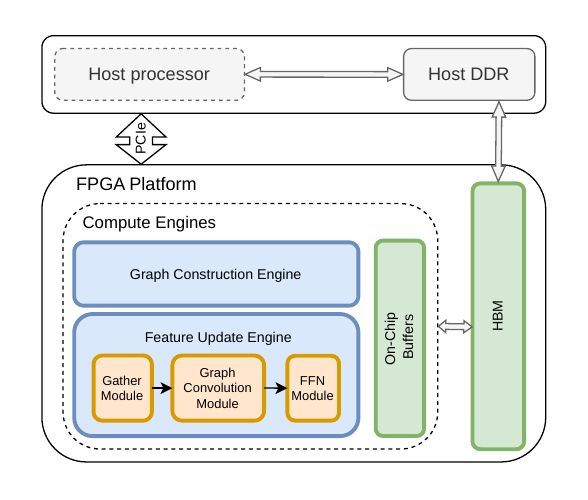}
\caption{Overall architecture of GraphLeap, featuring the Graph Construction Engine and Feature Update Engine and preprocessing and postprocessing modules. 
\label{fig:overview}}
\end{figure}

\subsection{Graph Construction Engine (GCE)}
The GCE computes a dilated $k$-NN graph by processing node features arranged as an $N \times D$ tensor. Features are tiled into blocks of $p_D = 32$ elements. A mesh of $p_N = 32$ distance processing elements (PEs) computes pairwise distances using the method defined in \ref{alg:dist}, where each PE processes $\lceil N/p_N \rceil$ nodes. Each PE contains $p_D$ parallel MAC units to compute distances across feature tiles. The GCE maintains min-heaps to extract the $k$ smallest distances and streams neighbor indices to the FUE via Edge Output FIFOs. 
\begin{figure*}[t]
\centering
\includegraphics[width=0.9\textwidth]{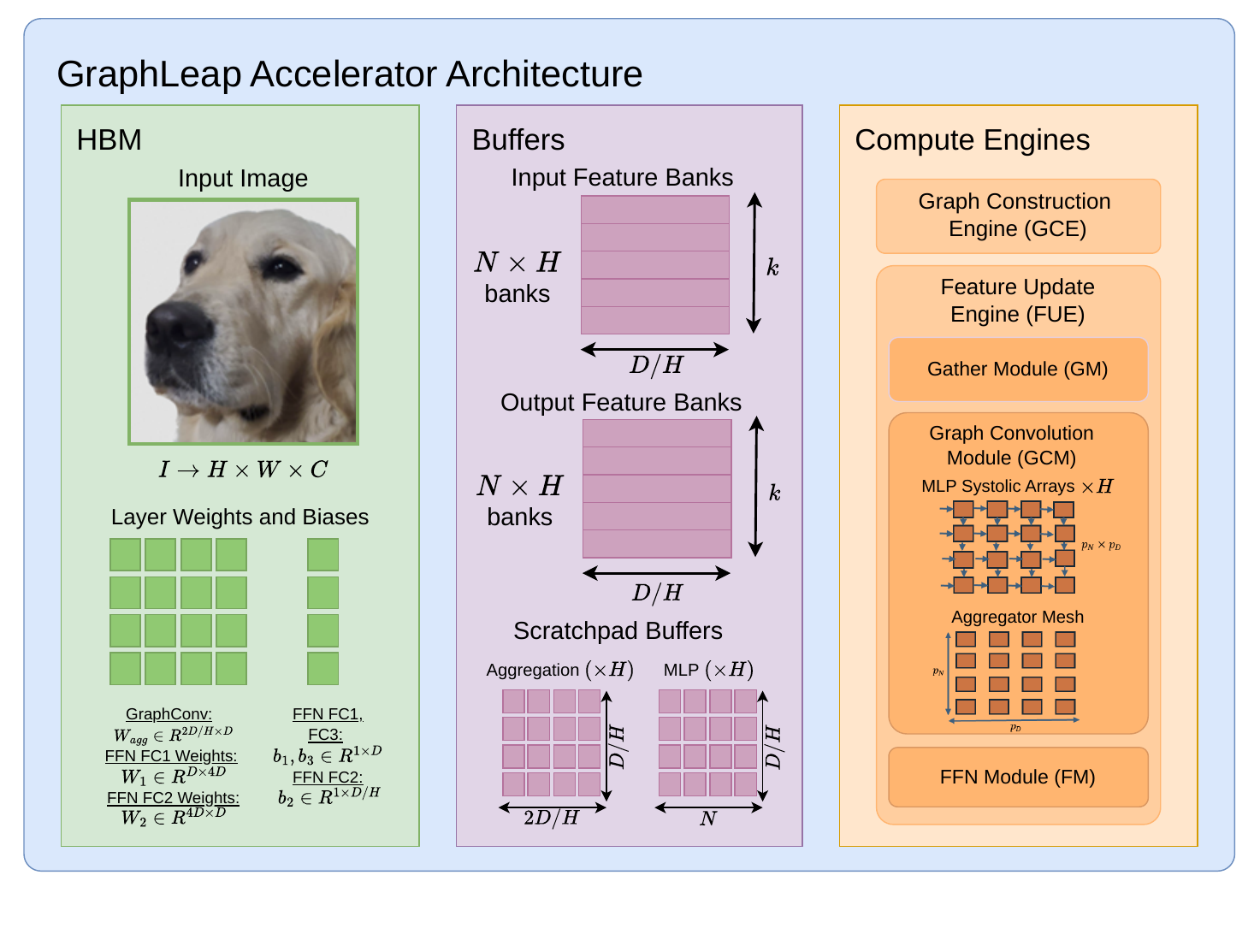}
\caption{Dataflow for the Graph Construction Engine and the Feature Update Engine, including the structure of the Graph Convolution Engine.
\label{fig:accelerator}}
\end{figure*}
\begin{algorithm}
    \caption{Tiled Distance Computation for GCM}
    \label{alg:dist}
    \begin{algorithmic}[1]
        \Statex \vspace{-2pt} 
        \For{\textbf{each} PE $p$ in parallel}
            \For{\textbf{each} node $i$}
                \For{\textbf{each} tile $t = 0$ to $\lceil D/32 \rceil - 1$}
                    \State Load $h_i^{(t)}, h_j^{(t)}$; $\text{p}[i][j] \mathrel{+}= ||h_i^{(t)} - h_j^{(t)}||_2^2$ \label{line:combined}
                \EndFor
                \State $d_{ij} = \sqrt{\text{p}[i][j]}$ for all $j$
            \EndFor
        \EndFor
        \Statex \vspace{-4pt} 
    \end{algorithmic}
\end{algorithm}

\subsection{Feature Update Engine (FUE)}
The FUE implements the logical \textbf{Grapher} and \textbf{FFN} blocks of the ViG model through specialized hardware modules: the Gather Module (GM), which implements GNN message passing; the Graph Convolution Module (GCM), which aggregates neighborhood information; and the Feedforward Neural Network (FFN) Module (FM), which transforms features and alleviates GNN oversmoothing.

\subsubsection{Gather Module (GM)}
The Gather Module (GM) serves as a high-speed data-acquisition front-end, bridging the asynchronous output of the GCE with the synchronous, high-throughput requirements of the Feature Update Engine (FUE). The primary challenge in ViG acceleration is the irregular memory access pattern during neighborhood gathering; the GM addresses this through aggressive prefetching and a multi-banked on-chip memory hierarchy. We employ an interleaved banking scheme where features are mapped to banks based on their token index $i \pmod H$. This ensures that spatially adjacent patches—which are frequently selected as neighbors in the early and middle layers of ViG architectures—reside in different physical banks, significantly reducing bank contention. When the GCE produces a cluster of $k$ neighbor indices for a target node, the GM's crossbar logic routes these requests to the appropriate banks, enabling the retrieval of high-dimensional feature vectors in a burst-streaming fashion.
\begin{figure}[b]
\centering
\includegraphics[width=0.48\textwidth]{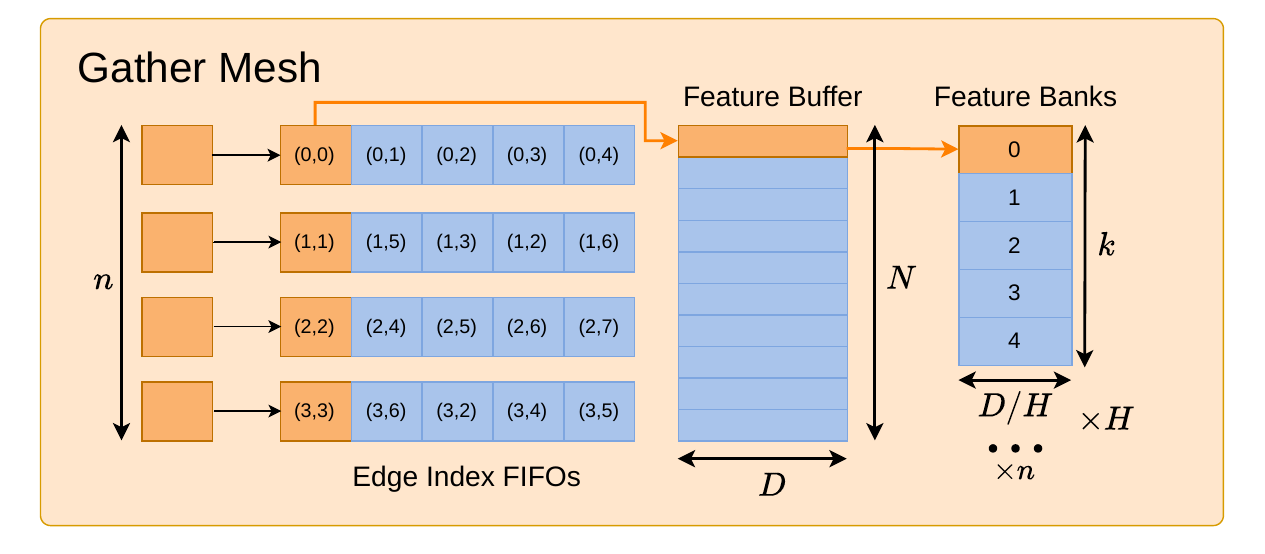}
\caption{The Gather Module, responsible for obtaining node and co-node features in preparation for graph convolution in the Feature Update Engine.
\label{fig:gather}}
\end{figure}
\subsubsection{Graph Convolution Module (GCM)}
The Graph Convolution Module (GCM) implements the Grapher’s message passing and linear transformations. Input and output projections are realized as standard fully connected layers with weights $W_{in}, W_{out} \in \mathbb{R}^{D \times D}$. For the central convolution, the aggregation weight matrix $W_{agg}$ is logically partitioned into $W_x$ and $W_m \in \mathbb{R}^{D \times D}$, allowing the systolic array to compute $y = x \cdot W_x + m \cdot W_m$ via fused accumulation without physically concatenating node features and messages. To support multi-head execution, $W_{agg}$ is further divided into $H$ heads, with each head operating on slices $W_{x,h}$ and $W_{m,h} \in \mathbb{R}^{\frac{D}{H} \times \frac{D}{H}}$ using a grouped dataflow that localizes data movement within systolic processing clusters.

\subsubsection{FFN Module (FM)}
The FM implements the ViG \textbf{FFN} block using a two-layer MLP ($D \rightarrow 4D \rightarrow D$). The fully connected layers $FC1$ ($W_1 \in \mathbb{R}^{D \times 4D}$) and $FC2$ ($W_2 \in \mathbb{R}^{4D \times D}$) are implemented as standard single-input datapaths without multi-head divisioning. The FM includes residual addition logic and a dedicated scratch buffer to compute $\text{FFN}(x) = \text{MLP}(x) + x$.

\subsection{Normalization and Activation Modules}

The accelerator integrates specialized normalization and activation units directly into the Feature Update Engine pipeline to enable on-the-fly processing and reduce HBM access. Layer Normalization computes per-node mean and variance across the feature dimension using parallel reduction and pipelined arithmetic, with gain and bias stored in local BRAMs, while BatchNorm2D applies channel-wise statistics in the convolutional stem using fully partitioned parameter buffers supporting $p_D$-way parallelism. Non-linear activation is provided by dedicated ReLU and GELU units, with ReLU implemented via a comparator-based clipper and GELU realized using a hardware-friendly PWL/LUT approximation similar to the approach taken by Sadeghi et al \cite{Peanovit}.

\subsection{Preprocessing and Postprocessing Modules}
The preprocessing and postprocessing pipeline bridges spatial image inputs and graph-based feature representations. A systolic-array–based convolutional stem converts raw pixels into feature maps, which are reshaped into $N$ node features by the patch embedding module with an optional linear projection using the shared MLP array. Positional context is incorporated via BRAM-based lookup tables indexed by 2D coordinates, and, in pyramidal ViG variants, node counts are reduced between stages using max pooling implemented with parallel reduction trees operating over $p_D$ dimensions.

\subsection{Pipelining Strategy}
The Grapher and FFN stages as defined in Section \ref{sec: ddvig} are implemented within the FUE and are pipelined with the GCE. We implement a dual-stage pipelining strategy that overlaps graph construction and per-layer weight prefetching with updating features. The GCE is scheduled to operate one layer ahead of the FUE. We employ Input and Output Feature Buffers (as seen in \ref{fig:accelerator} to store the features for the GCE and the FUE respectively and transfer the weights for layer $\ell + 1$ while the weights from layer $\ell$ are being used by the FUE.

\begin{figure}[h]
\centering
\includegraphics[width=0.48\textwidth]{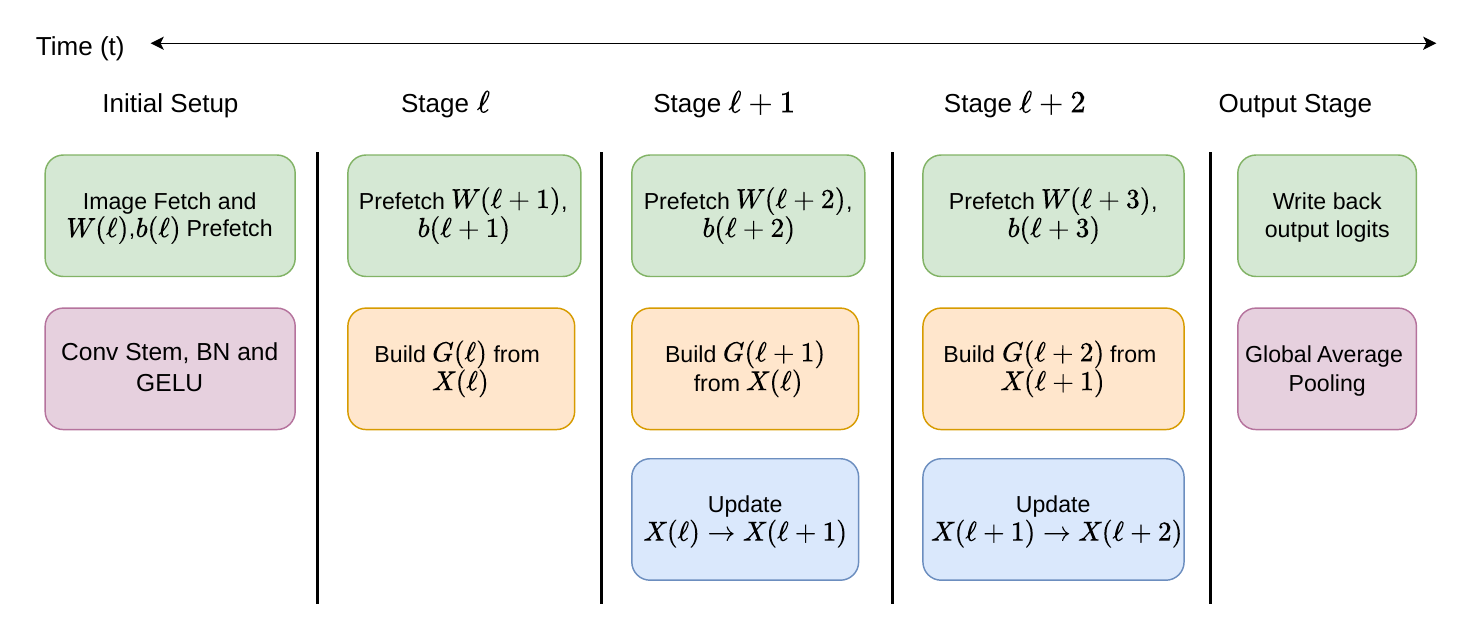}
\caption{Pipeline of the FPGA accelerator for end-to-end ViG inference, demonstrating how the tasks for the Graph Construction Engine and Feature Update Engine are scheduled for layers in Vision GNN.}
\label{fig:pipeline}
\end{figure}

\subsection{Resource and Performance Models}
The hardware cost of GraphLeap is dominated by parallel arithmetic units in the Graph Construction Engine (GCE) and the systolic-array–based Feature Update Engine (FUE). Compute resources (DSPs and LUTs) scale with node- and channel-level parallelism $p_N$ and $p_D$, and are modeled as $R_{\text{Total}} = (c_1 \Omega, c_2 \Omega)$, where $\Omega = p_N \cdot p_D$ and $c_1, c_2$ denote the per-MAC DSP and LUT costs. On-chip memory consumption $B_{\text{Total}}$ is driven by the look-ahead pipeline’s double buffering, requiring storage for $2(ND)$ node features and $Nk$ edge indices. We set $p_N = p_D = 32$ to saturate the Alveo U280’s DSP capacity.

Per-layer latency is determined by the slower of the overlapped engines, $T_{\text{layer}} = \max(T_{\text{GCE}}, T_{\text{FUE}}) + T_{\text{sync}}$. GCE latency is dominated by $k$-NN distance computation and sorting, $T_{\text{GCE}} \approx \frac{N}{p_N} \cdot \frac{N}{p_N} \cdot \frac{D}{p_D} + \frac{Nk}{p_N}$, but this quadratic cost is largely hidden by the look-ahead execution strategy, which overlaps graph construction for layer $\ell+1$ with feature updates for layer $\ell$.

FUE latency includes the Grapher and FFN stages. The Grapher latency, using partitioned $W_x$ and $W_m$ across $H$ heads, is $T_{\text{Grapher}} \approx 3 \left( \frac{N}{p_N} \cdot \frac{D}{p_D} \cdot D \right) + \left( \frac{N}{p_N} \cdot \frac{k}{H} \cdot \frac{D}{p_D} \right) + L_{\text{fused}}$, while the FFN stage incurs $T_{\text{FFN}} \approx \frac{8ND^2}{p_N p_D} + L_{\text{fused}}$. Here, $L_{\text{fused}}$ denotes the pipeline depth of the fused LayerNorm and activation units. Overall, parallelization along $p_N$ and $p_D$ significantly reduces linear-layer latency, keeping the FUE competitive with the high-throughput GCE.

\begin{table}[ht]
\centering
\scriptsize
\caption{Hardware Implementation Parameters on Alveo U280}
\label{tab:hw_parameters}
\renewcommand{\arraystretch}{1.1}
\setlength{\tabcolsep}{5pt}
\resizebox{\columnwidth}{!}{%
\begin{tabular}{lll lll}
\toprule
\textbf{Param.} & \textbf{Val.} & \textbf{Description} & \textbf{Param.} & \textbf{Val.} & \textbf{Description} \\
\midrule
$p_N$ & 32 & Parallel nodes (rows) & $BW_{HBM}$ & 460 GB/s & HBM2 Peak Bandwidth \\
$p_D$ & 32 & Parallel channels (cols) & $L_{\text{fused}}$ & 14 & Pipeline depth \\
$H$ & 16 & Parallel heads & $f_{\text{clk}}$ & 300 MHz & Target clock \\
$c_1$ & 1.0 & DSPs / MAC & $N_{buf}$ & 2 & Look-ahead buffer stages \\
\bottomrule
\end{tabular}}
\end{table}

\subsection{Implementation Details}

\subsubsection{Evaluated Models}
We evaluate GraphLeap on the Vision GNN (ViG) family, including Isotropic (Ti, S, B) and Pyramidal (Py-Ti, Py-S, Py-M, Py-B) variants. GraphLeap replaces dynamic graph construction with a look-ahead mechanism that decouples the Graph Construction Engine (GCE) from the Feature Update Engine (FUE), mitigating the primary latency bottleneck while preserving accuracy through fine-tuning.

\subsubsection{GraphLeap Reformulation and Fine-tuning}
The GraphLeap reformulation introduces an initial accuracy drop due to shifted connectivity. To recover performance, we fine-tune all models for 30 epochs on ImageNet-1K using AdamW with a cosine scheduler (base LR $=2\times10^{-5}$), recovering accuracy to within 1.4\% of original ViG Top-1 while enabling hardware parallelism.

\subsubsection{Hardware Implementation}
The accelerator is implemented on a Xilinx Alveo U280 FPGA using High-Level Synthesis (HLS). The GCE is placed on SLR 0 to leverage direct HBM2 access, while URAM-based look-ahead buffers overlap graph construction for layer~$\ell+1$ with feature updates for layer~$\ell$. A fixed-size design supports $224\times224$ and $448\times448$ inputs. Post place-and-route results show 300 MHz operation, with resource utilization reported in Table~\ref{table:fpga-resources}.
\begin{table}[h]
\centering
\caption{FPGA Resource Utilization (AMD-Xilinx Alveo U280)}
\label{table:fpga-resources}
\begin{tabular}{@{}lrrr@{}}
\toprule
Resource & Available & Used & Utilization \\ \midrule
LUT      & 1,182,240 & 798,200 & 67.5\%      \\
DSP      & 9,024     & 7,088   & 78.5\%      \\
BRAM     & 2,016     & 1,482   & 73.5\%      \\
URAM     & 960       & 860     & 89.6\%      \\ \bottomrule
\end{tabular}
\vspace{1pt}
\begin{flushleft}
\small \textit{*DSP utilization is intentionally high as our design targets full saturation of compute resources.}
\end{flushleft}
\end{table}
\section{Experimental Evaluation}
\label{sec: exp}

\subsection{Experimental Setup}
\subsubsection{Benchmark Dataset} 
We evaluate \textit{GraphLeap} using the ImageNet-1K \cite{deng2009imagenet} (ILSVRC 2012) dataset, consistent with the original ViG training protocol \cite{VisionGNN}. The dataset contains 1,000 object classes with 1.28 million training images and 50,000 validation images. All inputs are pre-processed to a $224\times224$ resolution.

\subsubsection{Model Architecture Specifications}
To evaluate the scalability and versatility of the \textit{GraphLeap} accelerator, we utilize both Isotropic and Pyramidal variants of the Vision GNN backbone. The architectural configurations, including the number of blocks, hidden dimensions ($D$), total parameters, and computational complexity (FLOPs), are summarized in Table~\ref{table:model_arch_specs}.

\begin{table}[ht]
\centering
\caption{Architectural Specifications of Vision GNN (ViG) Models \cite{VisionGNN}.}
\label{table:model_arch_specs}
\renewcommand{\arraystretch}{0.9}
\setlength{\tabcolsep}{4pt}
\begin{small}
\resizebox{\columnwidth}{!}{%
\begin{tabular}{@{}lccccc@{}}
\toprule
\textbf{Model} & \textbf{Structure} & \textbf{Blocks} & \textbf{Dim ($D$)} & \textbf{Params} & \textbf{FLOPs} \\ \midrule
ViG-Ti      & Isotropic & 12 & 192  & 0.7M  & 0.1G \\
ViG-S       & Isotropic & 12 & 320  & 2.4M  & 0.4G \\
ViG-B       & Isotropic & 16 & 640  & 10.5M & 2.1G \\ \midrule
ViG-Py-Ti   & Pyramidal & [2, 2, 6, 2] & [48, 96, 240, 384] & 2.9M  & 0.6G \\
ViG-Py-S    & Pyramidal & [2, 2, 6, 2] & [80, 160, 400, 640] & 8.2M  & 1.5G \\
ViG-Py-M    & Pyramidal & [2, 2, 16, 2] & [80, 160, 400, 640] & 18.0M & 3.4G \\
ViG-Py-B    & Pyramidal & [2, 2, 18, 2] & [96, 192, 480, 768] & 27.6M & 5.2G \\ \bottomrule
\end{tabular}%
}
\end{small}
\end{table}

\subsubsection{Baseline Platforms}
We benchmark \textit{GraphLeap} against CPU and GPU baselines utilizing the official Vision GNN implementation \cite{VisionGNNGithub}. Hardware specifications are detailed in Table \ref{table:specifications}.

\textbf{CPU Baseline:} An AMD EPYC 7763 processor is utilized with 64 physical cores ($T=64$) to ensure a fully multithreaded software baseline.

\textbf{GPU Baseline:} We utilize an NVIDIA RTX A5000 (Ampere architecture). Models execute via PyTorch v2.0.1 with a CUDA 12.2 backend.

\textbf{FPGA Implementation:} The proposed architecture is implemented on the AMD-Xilinx Alveo U280. The design is synthesized and implemented using the AMD-Xilinx Vitis v2025.2 flow at a target frequency of 300~MHz.

\begin{table}[ht]
\centering
\caption{Specifications of Evaluation Platforms}
\label{table:specifications}
\renewcommand{\arraystretch}{0.85}
\setlength{\tabcolsep}{4pt}
\begin{small} 
\resizebox{\columnwidth}{!}{%
\begin{tabular}{@{}lccc@{}}
\toprule
\textbf{Feature} & \textbf{CPU} & \textbf{GPU} & \textbf{FPGA (Ours)} \\ \midrule
Platform         & EPYC 7763    & RTX A5000    & Alveo U280           \\
Architecture     & 64C/128T     & Ampere        & Spatial HLS          \\
Frequency        & 2.45 GHz      & 1.17 GHz      & 300 MHz              \\
Peak Performance & 5 TFLOPS     & 27.8 TFLOPS  & 2.1 TFLOPS           \\
On-chip Memory   & 256 MB        & 24 GB        & 38 MB                \\
Ext. Memory BW   & 410 GB/s     & 768 GB/s     & 460 GB/s             \\ \bottomrule
\end{tabular}%
}
\end{small}
\end{table}

\subsubsection{Performance Metrics and Latency Definition} 
To ensure a fair and rigorous evaluation, the primary metric utilized is End-to-End (E2E) Latency for single-image inference ($\text{Batch}=1$). We define E2E latency ($L_{total}$) as the cumulative time elapsed from the host-to-device transfer of the input image to the final device-to-host transfer of the classification result. 

For our FPGA implementation, $L_{total}$ is defined as:
\begin{equation}
    L_{total} = T_{pcie\_in} + T_{kernel\_exec} + T_{pcie\_out}
\end{equation}
where $T_{kernel\_exec}$ encompasses the execution of all $N$ layers of the ViG model, including initial patch embedding, graph construction, GraphConv, and the final MLP head. Latency is captured via high-precision hardware cycle counters. For CPU and GPU baselines, this is measured using \texttt{torch.cuda.Event} synchronization, ensuring no asynchronous kernel overhead is omitted. While GPUs achieve peak throughput at large batch sizes, many real-world edge applications demand immediate response to streaming sensor data. Consequently, our evaluation prioritizes Batch=1 performance to quantify the true time-to-insight.

\subsection{GraphLeap Accuracy and Convergence}
The \textit{GraphLeap} reformulation decouples graph construction and feature updates. To recover the initial accuracy drop, we fine-tune for 30 epochs using AdamW with a cosine scheduler ($LR=2\times10^{-5}$), freezing the first 50\% of the backbone. As shown in Table~\ref{tab:accuracy_results}, this strategy allows ViG-Ti to recover to 73.28\% and ViG-Py-B to reach 82.34\%, matching original performance while enabling hardware-level parallelism.

\begin{table}[ht]
\centering
\caption{Top-1 Accuracy Analysis: Standard ViG vs. GraphLeap}
\vspace{-2mm}
\label{tab:accuracy_results}
\renewcommand{\arraystretch}{0.85}
\setlength{\tabcolsep}{5pt}
\begin{small} 
\resizebox{\columnwidth}{!}{%
\begin{tabular}{llcccc}
\toprule
\textbf{Structure} & \textbf{Model} &  \textbf{ViG} & \textbf{Pre-FT} & \textbf{Post-FT} & \textbf{Epochs}\\
\midrule
\multirow{3}{*}{Isotropic}  
& ViG-Ti & 74.5\% & 66.76\% & 73.28\% & 16 \\
& ViG-S  & 80.6\% & 69.24\% & 79.50\% & 9 \\
& ViG-B  & 82.6\% & 74.86\% & 80.66\% & 4 \\
\midrule
\multirow{4}{*}{Pyramidal}  
& ViG-Py-Ti  & 78.5\% & 76.66\% & 77.88\% & 6 \\
& ViG-Py-S   & 82.1\% & 79.47\% & 80.66\% & 6 \\
& ViG-Py-M   & 83.1\% & 79.54\% & 81.88\% & 6 \\
& ViG-Py-B   & 83.7\% & 81.44\% & 82.34\% & 3 \\
\bottomrule
\end{tabular}
}
\end{small}
\end{table}

\subsection{Accelerator Performance Analysis}
As shown in Table~\ref{tab:performance_final}, while the \textit{GraphLeap} reformulation yields modest algorithmic speedups on CPU and GPU platforms (1.03--1.23$\times$), these improvements are constrained by the underlying execution and memory hierarchies for standard Vision GNNs. The reformulation enables a different mapping on our FPGA accelerator, where graph construction and feature updates are executed in a tightly coupled, streaming pipeline. This eliminates intermediate data materialization and significantly reduces off-chip memory traffic, resulting in up to a 95.7$\times$ speedup over the multithreaded CPU baseline. 

\begin{table*}[h]
\centering
\caption{End-to-End Inference Latency (ms) and Speedup Analysis.}
\label{tab:performance_final}
\renewcommand{\arraystretch}{0.9} 
\setlength{\tabcolsep}{4pt} 
\begin{small} 
\resizebox{\textwidth}{!}{
\begin{tabular}{lc|cc|cc|cc}
\toprule
\multirow{2}{*}{\textbf{Model}} & \multirow{2}{*}{\textbf{Res.}} & \multicolumn{2}{c|}{\textbf{CPU (EPYC 7763)}} & \multicolumn{2}{c|}{\textbf{GPU (RTX A5000)}} & \multicolumn{2}{c}{\textbf{FPGA (U280)}} \\
& & ViG (ms) & GraphLeap (ms) & ViG (ms) & GraphLeap (ms) & ViG (ms) & GraphLeap (ms) \\
\midrule
\multirow{2}{*}{ViG-Ti} & $224 \times 224$ & 59.23 / 1.0$\times$ & 51.73 / 1.14$\times$ & 11.72 / 5.06$\times$ & 11.20 / 5.29$\times$ & 2.88 / 20.57$\times$ & 1.92 / 30.85$\times$ \\
                        & $448 \times 448$ & 640.9 / 1.0$\times$ & 610.4 / 1.05$\times$ & 78.8 / 8.13$\times$ & 75.0 / 8.55$\times$ & 46.9 / 13.67$\times$ & 31.3 / 20.48$\times$ \\
\midrule
\multirow{2}{*}{ViG-S}  & $224 \times 224$ & 108.16 / 1.0$\times$ & 105.38 / 1.03$\times$ & 16.63 / 6.50$\times$ & 16.41 / 6.59$\times$ & 2.94 / 36.79$\times$ & 1.96 / 55.18$\times$ \\
                        & $448 \times 448$ & 1340.1 / 1.0$\times$ & 1276.3 / 1.05$\times$ & 115.4 / 11.61$\times$ & 109.9 / 12.19$\times$ & 48.0 / 27.92$\times$ & 32.0 / 41.88$\times$ \\
\midrule
\multirow{2}{*}{ViG-B}  & $224 \times 224$ & 265.13 / 1.0$\times$ & 215.56 / 1.23$\times$ & 16.64 / 15.93$\times$ & 16.43 / 16.14$\times$ & 4.16 / 63.73$\times$ & 2.77 / 95.71$\times$ \\
                        & $448 \times 448$ & 3285.0 / 1.0$\times$ & 3128.5 / 1.05$\times$ & 115.6 / 28.42$\times$ & 110.1 / 29.84$\times$ & 67.8 / 48.45$\times$ & 45.2 / 72.68$\times$ \\
\midrule
\multirow{2}{*}{ViG-Py-Ti} & $224 \times 224$ & 78.93 / 1.0$\times$ & 68.42 / 1.15$\times$ & 13.23 / 5.97$\times$ & 13.13 / 6.01$\times$ & 2.88 / 27.41$\times$ & 1.92 / 41.11$\times$ \\
                           & $448 \times 448$ & 977.9 / 1.0$\times$ & 931.4 / 1.05$\times$ & 93.1 / 10.50$\times$ & 88.6 / 11.04$\times$ & 47.0 / 20.81$\times$ & 31.3 / 31.24$\times$ \\
\midrule
\multirow{2}{*}{ViG-Py-S}  & $224 \times 224$ & 113.21 / 1.0$\times$ & 102.03 / 1.11$\times$ & 13.47 / 8.40$\times$ & 13.18 / 8.59$\times$ & 4.68 / 24.19$\times$ & 3.12 / 36.29$\times$ \\
                           & $448 \times 448$ & 1402.7 / 1.0$\times$ & 1335.9 / 1.05$\times$ & 92.7 / 15.13$\times$ & 88.3 / 15.89$\times$ & 76.4 / 18.36$\times$ & 50.9 / 27.56$\times$ \\
\midrule
\multirow{2}{*}{ViG-Py-M}  & $224 \times 224$ & 187.53 / 1.0$\times$ & 186.77 / 1.00$\times$ & 23.15 / 8.10$\times$ & 23.05 / 8.14$\times$ & 5.85 / 32.06$\times$ & 3.90 / 48.08$\times$ \\
                           & $448 \times 448$ & 2314.1 / 1.0$\times$ & 2203.9 / 1.05$\times$ & 162.9 / 14.21$\times$ & 155.1 / 14.92$\times$ & 95.4 / 24.26$\times$ & 63.6 / 36.39$\times$ \\
\midrule
\multirow{2}{*}{ViG-Py-B}  & $224 \times 224$ & 314.20 / 1.0$\times$ & 280.20 / 1.12$\times$ & 25.11 / 12.51$\times$ & 24.89 / 12.62$\times$ & 6.66 / 47.18$\times$ & 4.44 / 70.77$\times$ \\
                           & $448 \times 448$ & 3892.9 / 1.0$\times$ & 3707.6 / 1.05$\times$ & 176.6 / 22.04$\times$ & 168.2 / 23.14$\times$ & 108.8 / 35.78$\times$ & 72.5 / 53.70$\times$ \\
\bottomrule
\end{tabular}
}
\end{small}
\end{table*}

\begin{figure*}
\centering
\includegraphics[width=0.9\textwidth]{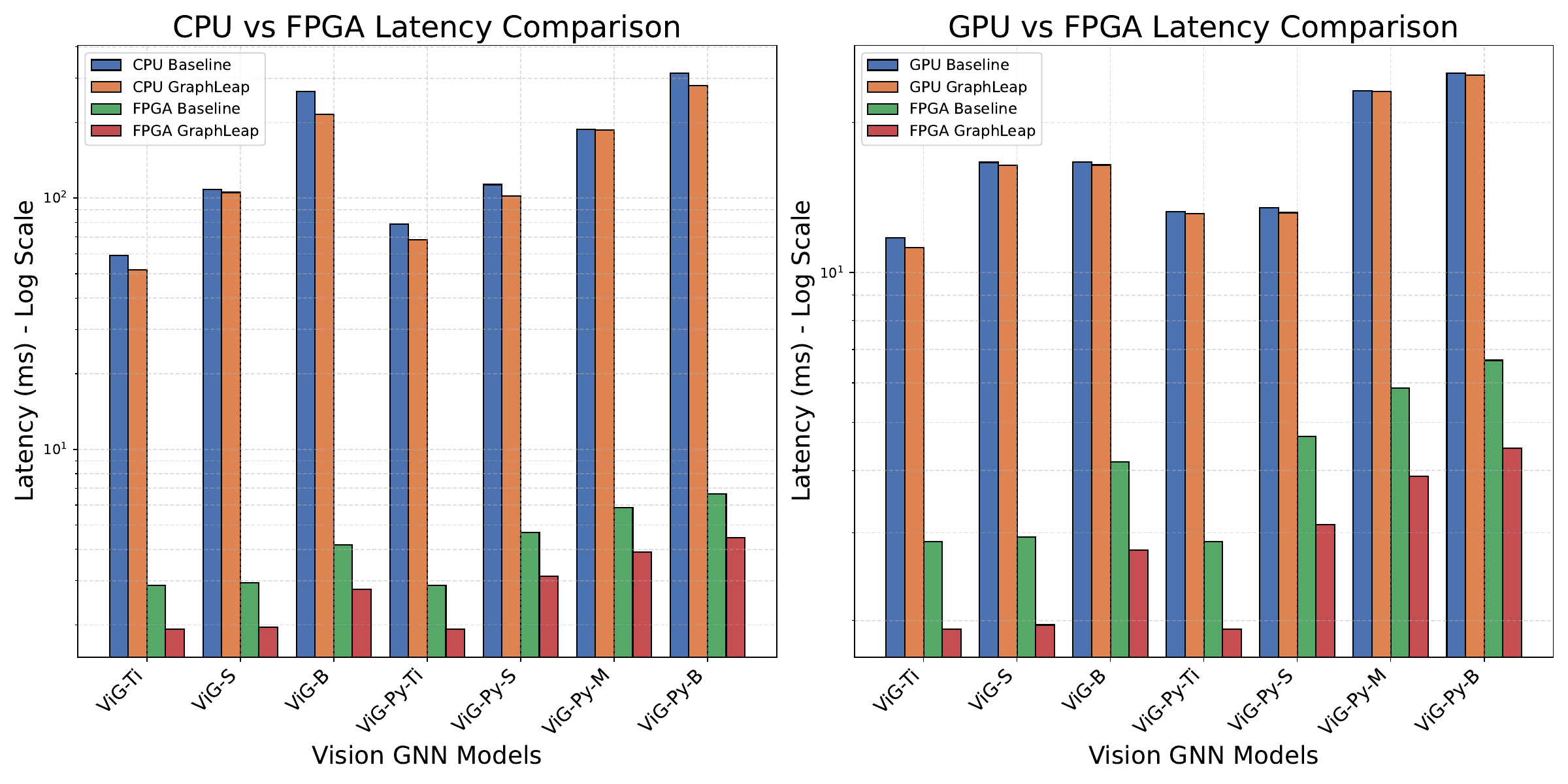}
\caption{End-to-End Inference Latency (ms) for ViG and GraphLeap variants (image resolution $224\times224$) across hardware platforms.
\label{fig:cpu_gpu_fpga}}
\end{figure*}

A visual comparison of end-to-end inference latency across the evaluated platforms is provided in Fig.~\ref{fig:cpu_gpu_fpga}. As demonstrated in Table~\ref{tab:performance_final}, while the \textit{GraphLeap} reformulation yields modest algorithmic speedups on CPU and GPU (ranging from 1.03--1.23$\times$), its deployment on our dedicated FPGA accelerator provides a substantial performance leap, achieving up to a 95.7$\times$ speedup over the multithreaded CPU baseline. 

This gap in performance stems from the mismatch between Vision GNN workloads and  throughput-oriented architectures. In particular, graph construction and neighborhood aggregation introduce fine-grained data dependencies, irregular memory accesses, and limited data reuse, which collectively hinder efficient parallelization on CPUs and GPUs. The GraphLeap reformulation defines a producer–consumer relationship between graph construction and feature updates, enabling these stages to be tightly coupled in hardware and executed in a streaming fashion. Rather than storing intermediary graph-structures in off-chip memory, edges and features are generated and consumed on-the-fly, significantly reducing memory traffic and eliminating redundant data movement.

\subsection{SOTA Comparison}
To evaluate our architecture, we compare \textit{GraphLeap} against recent high-performance FPGA accelerators for vision backbones, specifically DRViT~\cite{sun2025drvit} and UbiMoE~\cite{dong2025ubimoe}, as summarized in Table~\ref{table:sota_comparison}. While these works represent the state-of-the-art in accelerating Vision Transformers and Mixture-of-Experts models on similar AMD-Xilinx Alveo platforms, our design demonstrates the latency and accuracy achievable with similarly sized Vision GNNs.
\begin{table}[ht]
\centering
\caption{Comparison with SOTA FPGA Vision Accelerators}
\label{table:sota_comparison}
\renewcommand{\arraystretch}{0.85}
\setlength{\tabcolsep}{4pt}
\begin{small} 
\resizebox{\columnwidth}{!}{
\begin{tabular}{@{}lccc@{}}
\toprule
\textbf{Feature} & \textbf{DRViT \cite{sun2025drvit}} & \textbf{UbiMoE \cite{dong2025ubimoe}} & \textbf{GraphLeap (Ours)} \\ \midrule
Model Target & ViT-Base & M$^3$ViT (MoE) & ViG-B / Py-B \\
Platform & Alveo U250 & Alveo U280 & Alveo U280 \\
\midrule
Latency (ms) & $\sim$14.4$^{\dagger}$ & 13.6 & \textbf{2.77 -- 4.44} \\
Top-1 Accuracy & 81.8\% & 81.4\% & \textbf{80.66\% -- 82.34\%} \\ \bottomrule
\end{tabular}
}
\end{small}
\end{table}

As shown in Table~\ref{table:sota_comparison}, \textit{GraphLeap} achieves a latency of 2.77~ms (ViG-B), which is 5.19$\times$ faster than DRViT \cite{sun2025drvit} and 4.91$\times$ faster than UbiMoE \cite{dong2025ubimoe}. This is attributed to the total elimination of the $O(N^2)$ graph construction bottleneck via deep pipelining across hardware modules.
\section{Related Work}
\label{sec:related}
\subsection{Vision GNN Backbones}
Vision Graph Neural Networks (ViGs) model images as graphs over patch tokens with layer-wise dynamic $k$-NN construction, achieving competitive accuracy with lower compute than ViTs~\cite{VisionGNN,VisionTransformers,CNNs}. Variants enhance graph connectivity and aggregation: ViHGNN uses hypergraph neighborhoods~\cite{ViHGNN}, PVG and DVHGNN employ progressive or multi-scale dilated connectivity~\cite{PVG,DVHGNN}, and MobileViG/MobileViGv2 target edge deployment with sparse attention and lightweight convolutions~\cite{MobileViG,MobileViG2}. AdaptViG, SearchViG, and AttentionViG explore gated aggregation, architecture search, and attention-driven neighbor selection to further improve the accuracy--efficiency trade-off~\cite{AdaptiveViG,SearchViG,AttentionViG,RecentGNNVision}.

\subsection{Algorithmic Acceleration of Vision GNNs}
Algorithmic methods reduce the cost of dynamic graph construction and message passing in ViGs. GreedyViG limits $k$-NN search to axial directions~\cite{GreedyVIG}, ClusterViG builds sparse graphs over cluster centroids~\cite{ClusterViG}, and WiGNet uses windowed graph construction to exploit locality~\cite{WiGNet}. More generally, sparsification, sampling, quantization, binarization, and pruning have been explored in BiGCN, BitGNN, channel-pruned GNNs, and Early-Bird GCNs~\cite{BiGCN,BitGNN,ChannelPruningGNN,EarlyBirdGCN}, while hardware-aware search methods like MaGNAS optimize under deployment constraints~\cite{MaGNAS,GraphVisionNetworks}. Nonetheless, most approaches assume static graphs or coarse-grained sparsity and do not remove the per-layer dynamic $k$-NN overhead inherent to ViGs.

\subsection{FPGA Accelerators for GNNs and Vision Models}
Most FPGA GNN accelerators target static graphs and non-vision tasks. HyGCN mitigates irregular memory access via hybrid architectures and graph partitioning~\cite{HyGCN}, while GraphAGILE and GraphLily exploit streaming SpMV/SpMM pipelines on HBM-equipped FPGAs~\cite{GraphAGILE,GraphLily}. Other designs leverage mixed precision, near-memory processing, and hardware-friendly GNN reformulations, e.g., MEGA, PIM-GCN, Lift, Graphite, EvGNN and FTC-GNN~\cite{MEGA,PIM-GCN,Lift,Graphite,EvGNN,FTC-GNN}, with surveys noting a focus on sparse kernels and memory systems for static or slowly changing graphs~\cite{AComprehensiveSurveyGNNSurvey,SurveyGNNSpeedup,SurveyFPGAML,AccelerationAlgorithmsGNN}.  

For vision workloads, FPGA accelerators have focused on CNNs, ViTs, and state-space models. DRViT and UbiMoE co-design ViT blocks and MoE computation for efficiency~\cite{sun2025drvit,dong2025ubimoe}, while PEANO-ViT approximates non-linear ops to reduce hardware cost~\cite{Peanovit}. DIGC-FPGA accelerates only dynamic image graph construction in ViGs~\cite{DIGC_FPGA}. In contrast, our approach decouples graph construction from graph convolution across layers, enabling a streaming, pipelined ViG accelerator that overlaps $k$-NN construction with feature updates.

\section{Conclusion \& Future Work}
\label{sec:conclusion}
In this paper, we introduced \textit{GraphLeap}, a reformulation that decouples dynamic image graph construction from graph convolution across layers in Vision GNNs, and demonstrated how this look-ahead execution can be efficiently realized on FPGA. By exposing a streaming producer–consumer relationship between graph construction and feature updates, \textit{GraphLeap} enables a hardware implementation that eliminates intermediate data materialization, reduces off-chip memory traffic, and sustains fine-grained pipeline parallelism. Building on this reformulation, we developed a spatial dataflow accelerator that overlaps dilated k-NN graph construction with feature updates, avoids explicit copying of edge-wise concatenated messages, and achieves low-latency inference across both isotropic and pyramidal ViG models. Experimental results show that this co-design approach yields only modest gains on conventional CPU/GPU platforms, but translates into substantial end-to-end latency improvements on FPGA, with up to $95.7\times$ speedup over a multithreaded CPU baseline while maintaining competitive accuracy.

In future work, we plan to extend the proposed accelerator to support a broader range of graph construction schemes and graph convolution variants, enabling \textit{GraphLeap}-style designs to generalize across a wider class of Vision GNN architectures and further advance low-latency graph-based visual inference.

\section*{Acknowledgement}
This work is supported by the National Science Foundation (NSF) under grants CSSI-2311870 and OAC-2505107.
\bibliographystyle{IEEEtran}
\bibliography{main}

\end{document}